\documentclass[runningheads]{llncs}
\usepackage{graphicx}
\usepackage{amssymb}
\usepackage{amsmath}
\usepackage[misc]{ifsym}
\graphicspath{ {./} }
\newcommand{\netname}{\textsc{TimeHetNet}}

\newcommand{\iwata}{\textsc{HetNet}}

\usepackage{url}
\usepackage{booktabs}
\begin{document}


\title{Few-Shot Forecasting of Time-Series with Heterogeneous Channels}

\titlerunning{Meta-Learning for Time-Series with Heterogeneous Channels}

\author{Lukas Brinkmeyer$^*${\Letter}\orcidID{0000-0001-5754-1746} \and
Rafael Rego Drumond$^*${\Letter}\orcidID{0000-0002-6607-3208}\and
Johannes Burchert \and Lars Schmidt-Thieme \orcidID{0000-0001-5729-6023}}
\def\thefootnote{*}\footnotetext{Equal contribution}
\authorrunning{L. Brinkmeyer and R.R. Drumond et al.}

\institute{University of Hildesheim, Germany 
\email{$\{$brinkmeyer,radrumond,burchert,schmidt-thieme$\}$@ismll.uni-hildesheim.de} }

\tocauthor{Lukas~Brinkmeyer, Rafael~Rego~Drumond, Johannes~Burchert and Lars~Schmidt-Thieme}
\maketitle              \begin{abstract}

Learning complex time series forecasting models usually requires a large amount of data, as each model is trained from scratch for each task/data set.
Leveraging learning experience with similar datasets is a well-established technique for classification problems called few-shot classification. However, existing approaches cannot be applied to time-series forecasting because i) multivariate time-series datasets have different channels, and ii) forecasting is principally different from classification. In this paper, we formalize the problem of few-shot forecasting of time-series with heterogeneous channels for the first time. Extending recent work on heterogeneous attributes in vector data, we develop a model composed of permutation-invariant deep set-blocks which incorporate a temporal embedding. We assemble the first meta-dataset of 40 multivariate time-series datasets and show through experiments that our model provides a good generalization, outperforming baselines carried over from simpler scenarios that either fail to learn across tasks or miss temporal information.


\end{abstract} \section{Introduction}

\def\thefootnote{}\footnotetext{Accepted at ECML 2022}

Time-series research is a central area in the field of machine learning and is widely present in real-life problems and applications ranging from health to the financial sector \cite{lim2021time,krollner2010financial}, with time-series data being an essential modality in all of the industry. In particular, time-series forecasting has been in focus of research as it strives to forecast variables over a future time horizon which applies to most data currently being collected. Forecasts can be made for a complete horizon or just a single point in time. Forecasting on univariate time-series, meaning a signal that varies over time and comes without covariates that contain additional information, e.g., the day of the week, is a well-researched area, spanning decades of work with classical approaches being well-studied for all kinds of problem settings \cite{BoxArima,salinas2020deepar}. Recently, deep learning approaches are becoming more popular in this area, showing to outperform classical approaches when a sufficient amount of training data is available \cite{makridakis2018m4}. However, often this is not the case as many time-series datasets are limited in size, giving classical approaches the edge \cite{makridakis2018statistical}. Specifically, in the case of multivariate time-series data, this is a common problem because datasets have different sets of covariates, making it impossible to learn joint model attributes.
    Research for time-series shares a lot of commonality with research on image data since both areas are just special cases of structural data. As an example, both benefit from using convolutional networks and transformer-based approaches. However, one main difference in the respective state-of-the-art models is that the best approaches in computer vision rely almost exclusively on a deep feature extractor \cite{foret2020sharpness,tolstikhin2021mlp}. These models are pretrained on vast amounts of data pooled from various sources with a trend toward ever-larger collections (e.g. Imagenet \cite{deng2009imagenet}, JFT-300M \cite{hinton2015distilling}, JFT-3B \cite{chollet2017xception}) to facilitate the ever-growing models, which nowadays consist of several billion parameters \cite{dai2021coatnet}.
    Meanwhile, this is not easily possible on time-series data due to heterogeneous covariates. Thus, virtually all models are just trained on limited single tasks of time-series data. Looking at the state-of-the-art in the area of computer vision, our aim is to enable the training of a single model across a larger pool of various time-series datasets. The M4 competition was held with that objective \cite{makridakis2018m4}, by assembling a dataset consisting of 100.000 time-series from various datasets and domains, they analyzed the performance of forecasting approaches when applied to various tasks at once. The clear winner was a hybrid deep learning model by Smyl et al. \cite{smyl2020hybrid} outperforming any statistical model. However, the competition was limited to univariate time-series data to avoid the problem of dealing with heterogeneous channels and did not introduce unseen datasets in the test evaluation. Most real-world applications involve multivariate time-series data since a set of covariates is almost always given, which can aid forecasting greatly. These covariate channels can be, for example, additional sensors or simple information about the respective day and month.
    
    Learning a single model on a set of different tasks can be achieved through meta-learning.
    Meta-learning has been hugely successful in various areas of machine learning, with a special focus on computer vision and few-shot image recognition in particular. In contrast to classical machine learning, where a model is typically trained on a single dataset for one specific task, meta-learning aims at learning from a distribution of tasks which can vary in their target \cite{finn2017model,oh2021boil} or even their predictors \cite{iwata2020meta,brinkmeyer2019chameleon}.
    Meta-learning techniques have been successfully applied to various areas of machine learning including few-shot classification \cite{hou2019cross}, hyperparameter optimization \cite{feurer2015initializing,jawed2021multi}, reinforcement learning \cite{gupta2018meta} and neural architecture search \cite{liu2018darts}.
    In particular, research in few-shot learning has seen an immense rise in popularity, with methods undergoing fundamental changes and benchmarks significantly improving over a very short period of time. Motivated by the fact that humans require only a few examples to correctly classify previously unseen objects based on their past experience, few-shot learning strives to learn models which can generalize to novel tasks based on task-agnostic information extracted from a large set of tasks.
    Most meta-learning approaches still require a homogeneous representation across tasks, rendering them not feasible in the application to multivariate time-series tasks with heterogeneous channels. In recent works, various approaches were published to enable machine learning on sets by introducing permutation-invariant and equivariant layers \cite{zaheer2017deep,maron2018invariant}. The work of Iwata et al. \cite{iwata2020meta} incorporated these permutation-invariant layers in a few-shot learning approach to enable learning on vector data with heterogeneous attributes. Encouraged by these findings, we propose the first model for few-shot forecasting on time-series tasks with heterogeneous channels.
Our main contributions are as follows:
    \begin{enumerate}
        \item We formalize the problem of few-shot forecasting of time-series with heterogeneous channels for the first time.
        \item We develop a model for this new problem that extends prior work on vector data in a principled way.
        \item We assemble the first meta-dataset of 40 multivariate time-series datasets and thus provide a public benchmark for future research.
        \item We show that our model provides a good generalization, outperforming baselines carried over from simpler scenarios that either fail to learn across tasks or miss temporal information.
    \end{enumerate}

 \section{Related work}
 
    There exists a vast amount of research on time-series forecasting and few-shot learning in literature, but the intersection of the two is still very limited, with no existing approaches dealing with few-shot forecasting for time-series tasks with heterogeneous channels. 
    In this section, we will discuss the research of these related areas in a concise way and point out the most important distinctions.

\textbf{Time-series forecasting} focuses on identifying temporal patterns in given data. Historically this was done with methods like ARIMA \cite{BoxArima}. In the field of machine learning, CNN and RNN architectures \cite{salinas2020deepar,RangapuramDeepState} were used to significantly outperform these methods. A further improvement came with the incorporation of attention layers \cite{vaswani2017attention} in time-series models \cite{qin2017dual}. While proving to be very effective, the quadratic complexity of attention comes with a high computational cost. Recent architectures like the Reformer \cite{kitaev2020reformer}, Yformer \cite{madhusudhanan2021yformer}, and Informer \cite{zhou2021informer} focused on reducing this cost by introducing restricted attention layers to effectively approximate the full attention mechanism. Currently, the best performing model architectures are SCINet \cite{liu2021time} and N-BEATS \cite{nbeats} on all common datasets and we will compare against them as baselines. 
    
In relation to the problem setting in this work, learning across time-series stemming from different datasets has also been the goal of the popular M4 time-series forecasting competition \cite{makridakis2018m4}. However, it was limited to univariate time-series and, more importantly, designed as a classical forecasting problem and not a meta-learning one, meaning that the test set only contained future windows of datasets seen during training. In contrast, our work aims at generalizing to a new time-series dataset during testing, which renders the winning approaches of the M4 competition not applicable to our problem. 
\textbf{Few-Shot Learning} describes a subarea of meta-learning that deals with evaluating tasks of unseen classes or even datasets with very few labeled samples \cite{wang2020generalizing}. By learning from a large collection of related tasks, the model is trained to capture task-agnostic knowledge, which can then be used for a fast adaptation to a novel task that shares this similarity. Different approaches have been proposed with this goal in mind. Gradient-based methods rely on second-order gradient information that is passed across tasks to optimize meta-parameters \cite{finn2017model,nichol2018first,rusu2018meta}. Neighbor-based approaches learn a metric embedding space to compare novel tasks \cite{vinyals2016matching,snell2017prototypical} while memory-based approaches rely on recurrent components to memorize a representation of the previous tasks \cite{santoro2016meta,munkhdalai2017meta}.
    
 All these methods, however, require a homogeneous predictor and target space in order to learn a joint distribution. 
 One of the first methods to attempt few-shot learning on homogeneous predictors was \textsc{chameleon} \cite{brinkmeyer2019chameleon}, which used a convolutional encoder to align tasks from similar domains to a common attribute space before utilizing gradient-based few-shot methods. Similarly, other works tried to learn across tasks with varied label spaces \cite{doi:10.1137/1.9781611976236.45,oh2021boil}. 
 Finally, Iwata et al. \cite{iwata2020meta} proposed a model that uses deep sets \cite{zaheer2017deep} based blocks to compute a task-embedding over predictor and targets of training samples (support data), which then can be combined with new unlabeled samples (query data) to perform classification or regression without the need of retraining or fine-tuning, similar to neighbor-based approaches (we will refer to this method as \iwata{} throughout the rest of the paper). The main advantage of \iwata{} is that, since it uses the deep sets formalization, it is invariant to the order of attributes and samples in both query and support set. So far, these approaches are limited to simple vector data and not applicable to structural data.\\

\textbf{Few-shot learning for Time-series data}. Few works have been published that apply few-shot learning or even meta-learning to time-series data.
We argue that this is due to the fact that, in contrast to image data, it is not readily possible to learn a single feature extractor across tasks stemming from different datasets when dealing with multivariate time-series data. Thus, published approaches can be divided into two groups. The first group of approaches utilizes meta-learning techniques to train across slices of the same time-series with homogeneous channels. This includes approaches that combine classical time-series regression with gradient-based meta-learning \cite{arango2021multimodal} and approaches that utilize metric-based meta-learning in combination with shapelet learning \cite{tang2020interpretable}. These methods are not applicable to our problem setting as they are not equipped to learn across tasks with heterogeneous channels.

Second, there are meta-learning approaches for time-series data that limit their problem setting to univariate time-series tasks in order to learn a single feature extractor without having to deal with heterogeneous channels.
Iwata et al. \cite{iwata2020few} proposed a method to embed tasks through BiLSTM and regular LSTM layers. Narwariya et al. \cite{narwariya2020meta} utilized Resnet to embed each time-series to a vector, and then trained across tasks with \textsc{Reptile} \cite{nichol2018first} on a meta-dataset of 41 univariate UCR datasets. Lastly, Oreshkin et al. \cite{oreshkin2020meta} showed how N-BEATS can be used for zero-shot time-series forecasting by rephrasing it in a meta-learning formalization. None of these approaches are capable of dealing with multivariate time-series tasks with heterogeneous channels, which is the focus of this work. Nevertheless, we compare our approach against N-BEATS for zero-shot time-series forecasting by using only the target channel to show that incorporating covariate channels is absolutely necessary for this problem setup.
Our method called \netname{} serves as the first few-shot time-series forecasting model for multivariate time-series data that can learn across different tasks with heterogeneous channels. 

\section{Methodology}
In this work, we want to propose the first work in the intersection of few-shot learning and multivariate time-series forecasting. We will first formalize the problem of time-series forecasting on a single task before extending it to a few-shot learning setting across a meta-dataset of tasks with heterogeneous channels.
    
\subsection{Problem setting}
In the \textbf{(vanilla) time-series forecasting problem}, a time-series $x$ with $C$ channels is a finite sequence of vectors in $\mathbb{R}^{C}$. Their space is denoted by $\mathbb{R}^{* \times C}:=\bigcup_{T\in\mathbb{N}}\mathbb{R}^{T\times C}$ with time-series length $|x|:=T$. Time-series forecasting data with a single univariate target time-series and $C$ predictor channels is then given by:
\begin{equation}
\mathcal{D}:=\{(x_{1},y_{1},x'_{1},y'_{1}),...,(x_{N},y_{N},x'_{N},y'_{N})\} \in \mathcal{X}\times\mathcal{Y}
\end{equation}
with $x,x'\in\mathbb{R}^{*\times C}$ and $y,y'\in\mathbb{R}^{*}$, sampled from an unknown distribution $p$, where $x,y$ are predictors and targets up to a reference time point $t_{0}-1$ and $x',y'$ denote the corresponding future predictors and targets starting from time point $t_{0}$ up to $T$. The predictors can also be described as future covariate information for the target. Given a loss function ${l:\mathbb{R}^{*}\times\mathbb{R}^{*}\rightarrow \mathbb{R}}$, we want to learn a function $\hat{y}:\mathbb{R}^{*\times C}\times\mathbb{R}^*\times \mathbb{R}^{*\times C} \rightarrow \mathbb{R}^{*}$ called model with minimal expected loss over the data:
\begin{equation}
    \mathbb{E}_{(x,y,x',y')\sim p} \quad l(y',\hat{y}(x,y,x'))
\end{equation}
Extending this formalization, the problem of \textbf{few-shot time-series forecasting across tasks with heterogeneous channels} is then given by a sample $D:=\{(\mathcal{D}_1^{s},\mathcal{D}_1^{q}),...,(\mathcal{D}_{m}^{s},\mathcal{D}_{m}^{q})\}$ called meta-dataset of pairs $\mathcal{D}^{s},\mathcal{D}^{q}\in (\mathcal{X}\times\mathcal{Y})$ from an unknown distribution $p_{m}$ of dataset pairs, and a function ${\mathcal{L}:\mathcal{Y}\times\mathcal{Y}\rightarrow\mathbb{R}}$. Each pair is called a task and consists of support data $D^{s}$ for which the full instance $(x,y,x',y')$ is known during prediction time and the query data $D^{q}$ for which the future target $y'$ is not known during prediction time. The number of predictor channels $C$ varies across tasks between $C_{\min}$ and $C_{\max}$. 
In few-shot learning, the number of samples $N^{s}$ in the support data $D^{s}$ is typically low.
We want to find a function $\hat{y}:\mathcal{X}\times(\mathcal{X}\times\mathcal{Y})^{*}\rightarrow \mathcal{Y}$ called meta-model with minimal expected loss:
\begin{equation}
    \mathbb{E}_{(\mathcal{D}^{s},\mathcal{D}^{q})\sim p_{m}} \frac{1}{|\mathcal{D}^{q}|} \sum_{(x,y,x',y')\in \mathcal{D}^{q}} \mathcal{L}(y',\hat{y}(x,y,x',\mathcal{D}^{s}))
\end{equation}
In this work, the loss $\mathcal{L}$ is chosen to be the mean squared error (MSE) averaged over the query datasets of our meta-dataset.
By weighting only a single time step of the target $y'$, we can learn a point-forecasting model with a specific focus on that one point. 

\subsection{Model formulation}

In this work, we extend \iwata{} \cite{iwata2020meta} which is a permutation-invariant model for few-shot classification on vector data with heterogeneous attributes. Their model relies on nested deep set-blocks \cite{zaheer2017deep} which were first published as a means to enable machine learning models to process sets by introducing a family of permutation-invariant functions. Each deep set-block consists of an inner function $f$ that is applied on each element of the set and an outer function $g$ which is applied after aggregating the output of $f$ such that the block is permutation-invariant to the order of elements. The architecture consists of an inference network that extracts latent task-dependent features of the support data $D^{s}$, which are then used by the prediction network in conjunction with the query data $(x,y,x')\in D^{q}$ to generate the forecast for the future targets $y'$ of a given task. \\

\begin{figure}[t!] \label{network} \centering
    \includegraphics[width=\textwidth]{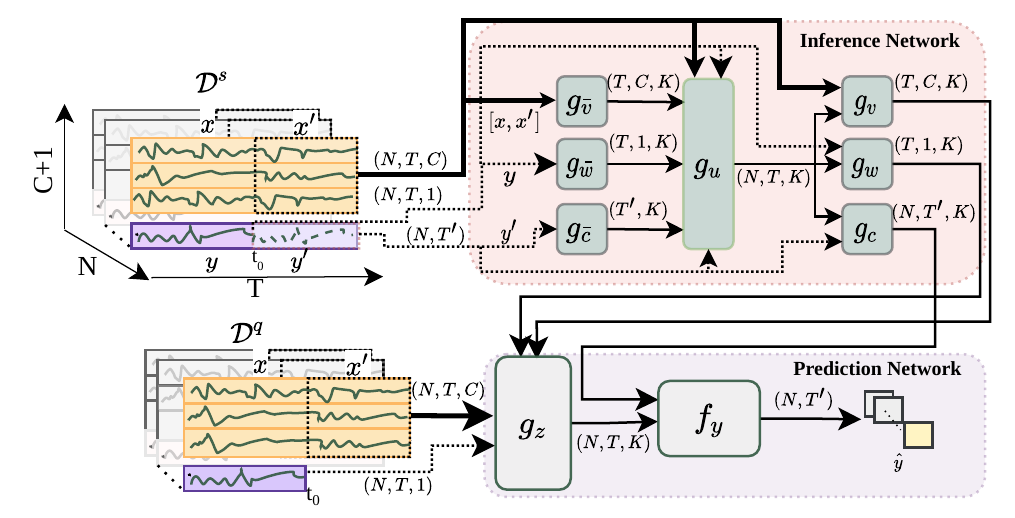}
    \caption{\netname{} Architecture. Values in parenthesis represent the output shape of each layer. $x$ and $y$ represent the predictor and target channels respectively, while $x^{\prime}$ and $y^{\prime}$ represent future predictor and targets. $N$ represents the number of samples in the set, $T$ represents the maximum time length, $C$ describes the number of channels/attributes in each sample. $K$ represents the latent space embedding. $T^{\prime}$ represents the number of future points of $y^{\prime}$ we want to predict. For readability: Raw predictors are bold, raw targets are dotted, and latent tensors are regular arrows.}
\end{figure}

\textbf{Inference Network}. We adapted the formalization of \iwata{} \cite{iwata2020meta} to our proposed problem setting by extending it from simple vector data to a forecasting task on structural data. A conceptual depiction of our model is shown in Figure~\ref{network}. First, using the support data $D^{s}$ we compute the target embeddings $\bar{w}$ and $\bar{c}$ for $y$ and $y'$ respectively, and a predictor embedding $\bar{v_{i}}$ for each predictor channel $i\in C$ by aggregating across the instances such that the block is permutation-invariant to their order:

        \begin{equation} \label{bar}
        \bar{v}_{i} = g_{\bar{v}} \left( \frac{1}{N} \sum_{n=1}^{N}  f_{\bar{v}}([x_{ni},x'_{ni}])\right) \quad \forall i\in C
        \end{equation}
        \begin{equation*}
        \bar{w} = g_{\bar{w}} \left( \frac{1}{N} \sum_{n=1}^{N} f_{\bar{w}}(y_{n}) \right), \quad
        \bar{c} = g_{\bar{c}} \left( \frac{1}{N} \sum_{n=1}^{N} f_{\bar{c}}(y'_{n}) \right)
        \end{equation*}
Here, $\bar{v}\in \mathbb{R}^{C\times T\times K}$ with $K$ being the latent output dimension of $g_{\bar{v}}$, $T$ the length of the time-series task at hand, $N$ the number of samples in the support data $D^s$ and $[\cdot,\cdot]$ a concatenation.
Through concatenation of the embeddings with the respective support data followed by another deep-set block, we can generate an embedding for each instance in the support data $D^{s}$ through aggregation over the predictor channels:
                \begin{equation}\label{ulayer}
                    u_{n} = g_{u} \left( \frac{1}{C} \sum_{i=1}^C \Big(f_{u}([x_{ni},x'_{ni},\bar{v}_{i}])\Big) + f_{o}([y_{n},\bar{w}]) + f_{p}([y'_{n},\bar{c}])\right) \quad \forall n\in N
                \end{equation}
        where $C$ is the number of predictor channels of the task and $u\in\mathbb{R}^{N\times T\times K}$.
        By concatenating these instance-wise embeddings with the respective support data before repeating the block structure of Equation \ref{bar}, we can again generate the predictor and target embeddings, only this time they are computed with regard to the entire support set. The embeddings for the future target $y'$ are not aggregated over the number of instances, as they are fed directly into the final network $f_y$ generating the forecast on an instance-level:
                \begin{equation} \label{vlayer}
                v_{i} = g_{v} \left( \frac{1}{N} \sum_{n=1}^{N}  f_{v}([x_{ni},{x'_{ni}},u_{n}])\right) \quad \forall i\in C
                \end{equation}
                
                \begin{equation*}
                w = g_{w} \left( \frac{1}{N} \sum_{n=1}^{N} f_{w}([y_{n},u_{n}]) \right), \quad
                c = f_{c}([y'_{n},u_{n}])
                \end{equation*}
                  
        \textbf{Prediction Network}. The embeddings for the predictors $v$ and the past targets $w$ are concatenated with the predictors $x$, $x'$, and past target $y$ of the query data $D^{q}$. This concatenation is passed to a deep set-block consisting of networks $g_z$ and $f_z$ which compute the per-instance features:
                \begin{equation}\label{zlayer}
                        z_{n} = g_{z} \left( \frac{1}{C} \sum_{i=1}^{I} f_{z}([x_{ni},y_{n},x'_{ni},v_{i},w]) \right) \quad \forall n\in N^{q}
                \end{equation}
        where $N^q$ is the number of samples in $D^q$. Finally, the prediction for a query sample $n$ is made by passing the embedding $z$ of $D^{q}$ and the embedding $c$ of the future target of $D^{s}$ to a final network $f_y$:
                \begin{equation}
                        \hat{y}_{n} = f_{y}([z_{n},c])
                \end{equation}
                
        The full \netname{} then contains the following neural networks $g_{\bar{v}}$, $f_{\bar{v}}$, $g_{\bar{w}}$, $f_{\bar{w}}$, $g_{\bar{c}}$, $f_{\bar{c}}$, $g_{u}$, $f_{u}$, $f_{o}$, $f_{p}$, $g_{v}$, $f_{v}$, $g_{w}$, $f_{w}$, $f_{c}$, $g_{z}$, $f_{z}$. We share the parameters between the network pairs $(f_{\hat{v}},f_{\hat{w}}),(g_{\hat{v}},g_{\hat{w}}),(f_v,f_w),(g_v,g_w)$.

 \section{Experimental setup}  
    
    In order to evaluate \netname{}, we have performed an extensive evaluation by creating a meta-dataset to learn few-shot forecasting on multivariate time-series data and comparing the performance of our method to related baseline methods. In this section, we will describe our meta-dataset construction, our baseline methods, the experimental procedures with observed results, and finally, a discussion of our findings.
    
    \begin{figure}[t!] \label{result1} \centering
        \includegraphics[width=0.78\textwidth]{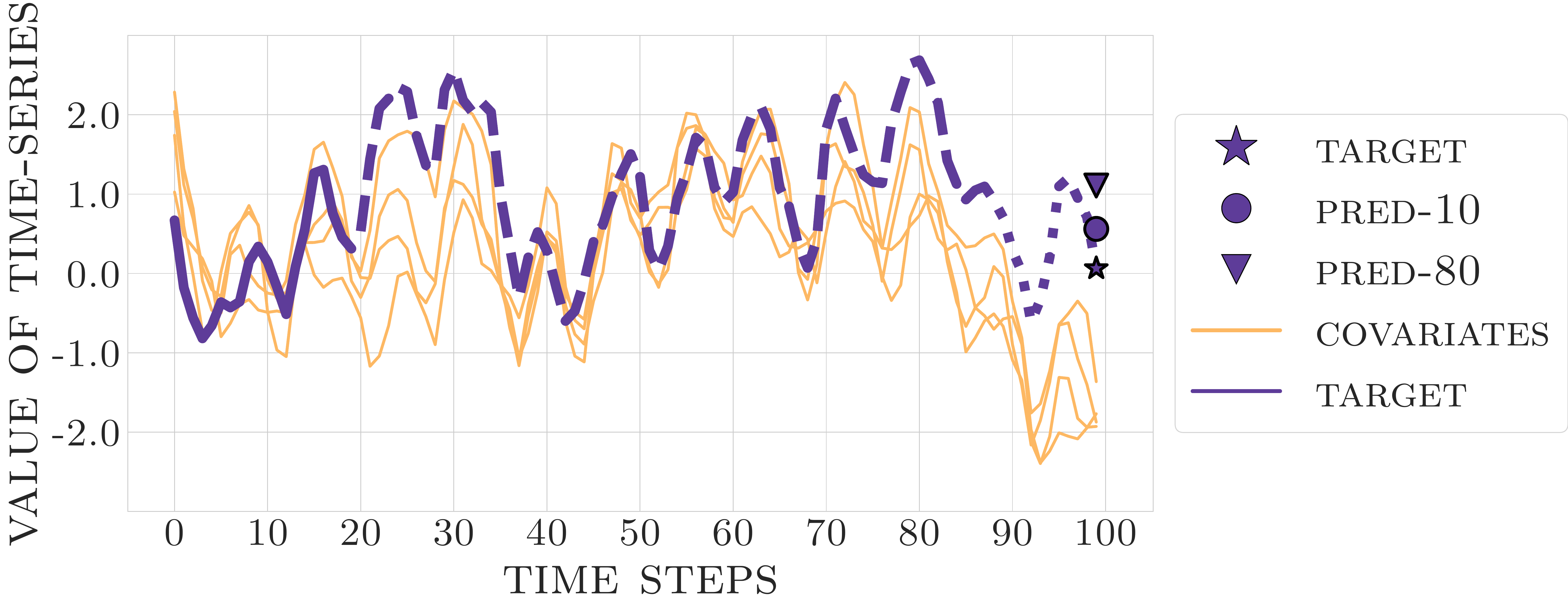}
        \caption{\textbf{Task visualization with forecast:} The figure shows the four covariate channels (orange) and the target channel (purple) of a query instance for a task sampled from \textit{HandMovementDirection}, and the predictions for the two models trained to predict $t_{0}+10$ (star) and  $t_{0}+80$ (triangle) which predict $10$ and $80$ time steps ahead respectively. The solid target line represents the $20$ target time steps $y$ which serve as input to the model trained to predict $t_{0}+80$, while the model trained to predict $t_{0}+10$ receives the target data including the dashed line mark. The dotted line shows the target over the last $10$ steps.}
    \end{figure}

    \subsection{Meta-dataset}
        In order to evaluate our approach, we assembled the first meta-dataset for few-shot forecasting on multivariate time-series data. For that purpose, we collected 40 multivariate time-series datasets consisting of popular forecasting datasets like the \textit{ETT} and \textit{ECL} \cite{zhou2021informer}, as well as datasets from the  \textit{Monash} time series forecasting archive, \cite{godahewa2021monash}, 3 \textit{Kaggle} datasets \cite{cnc,mining,plant}, the \textit{UCR} and \textit{UEA} time series archive \cite{bagnall2018uea,dau2019ucr}, and also the sparse motion capture dataset \textit{PeekDB} \cite{drumond2018peek}.
The details on all datasets are summarized in Table 3 (supplementary material). A single task is sampled from a dataset by randomly selecting between $C_{\min}$ and $C_{\max}$ of the channels, a random slice of the temporal dimension of size $T$ and $N_{Q}$+$N_{S}$ samples for the query and support split of the task. We normalized the channel of each task to mean zero and standard deviation one. In our experiments, we set the number of query and support instances to 20 each, the time length to $100$, and varied the number of channels between $5$ and $10$. In case a dataset has only a single instance, we sample multiple temporal slices for a single task. Furthermore, one channel is selected to be the target channel, with the remaining channels serving as covariate information (during sampling, we make sure that the target channel is not included in the covariates). Finally, the last time step $t=100$ of the target channel for a given instance is chosen as the future target $y'$, while the last $p$ time steps of the target channel $y$ are removed, thus creating a multivariate time-series forecasting task with covariate information with the aim of forecasting $p$ steps ahead. We evaluated our approach for $p=1, 10, 80$ and $100$, where $p=1$ corresponds to forecasting the next time step $t+1$ with the target channel $y$ including the first $99$ steps, and $p=80$ to forecasting the time step $t+80$ with the target channel including the first $20$ time steps. The experiments with $p=100$ demonstrate an uncontrolled scenario where only covariate channels are given in addition to the future target $y'$ for the support instances. A visualization of a single task is given in Figure \ref{result1} including the forecast of our approach.

    \subsection{Experimental details}
    
        We conducted a 5-fold cross-validation with each fold having 8 datasets in meta-test, 8 in meta-validation, and 24 in meta-training. During each epoch of meta-training, we sample $10$ meta-batches, where each meta-batch includes one task per dataset: 24 tasks in meta-training and 8 in meta-validation. For the sake of comparability, we generated 11.000 tasks for each dataset in meta-testing beforehand (1.000 tasks per channel size 5 to 10). The final meta-test performance can then be computed by evaluating the model on each of the 440.000 tasks in the fixed test set while guaranteeing comparability between different models. Instead of evaluating our model for the average loss over all time steps, we evaluate the model for individual time steps. By doing this, we want to emphasize the concrete performance differences for close and far events without overlap.
        For our approach \netname{}, all networks $f$ and $g$ in the deep set-blocks consist of three layers. In all configurations, $f_{y}$ is a feed-forward network. After optimizing the concrete architecture on the validation tasks, we selected GRU layers for the deep-set blocks formalized in equations \ref{bar} and \ref{zlayer} and convolutional layers for \ref{ulayer} and \ref{vlayer}. All hyperparameters for all approaches were optimized via grid search. Our model was trained with Adam \cite{kingma2014adam} for a maximum number of 15.000 epochs with early stopping over the validation tasks.
        More details on our experimental setup and the concrete hyperparameters can be found in our supplementary material.

        \begin{table}[t!] 
        
        \centering
        \caption{Experimental Results across all folds. All scores represent Mean Squared Error. Standard deviation is computed over 5 repeated experimental runs. Oracle Channel gives the best possible performance if the best control channel is known. Bold-faced results represent the best scores.}
        \label{resultsmain}
        \begin{tabular}{cc|cccc}
            \hline
               Category & Method &  ~~~$t_0+1$~~ & ~~~$t_0+10$~~~ & ~~~$t_0+80$~~~ & ~~$t_0+100$~~ \\ \hline\hline
Proposed & \netname{} & \textbf{0.148}       & \textbf{0.389}       & \textbf{0.509}       & \textbf{0.579} \\ 
                        & (ours)    &  $\pm$~ \textbf{0.003}  &  $\pm$~ \textbf{0.007} &  $\pm$~ \textbf{0.006} &  $\pm$~ \textbf{0.004}\\ \hline
                Meta-Learning & \iwata \cite{iwata2020meta}  & 0.178      & 0.413       & 0.524       & 0.582  \\
                              &  &  $\pm$~ 0.002 &  $\pm$~ 0.003 &  $\pm$~ 0.003 & $\pm$~ 0.006 \\ \hline
               & \textsc{zero prediction} &  1.006 & 1.006 & 1.006 & 1.006\\
            Heuristic   & \textsc{last time step}  &  0.215 & 0.899 & 1.404 & $\times$\\
               & ~~\textsc{avg time step}~~  &  0.867 & 0.867 & 0.867 & 0.867\\  \hline & \textsc{GRU} & 0.531 & 0.692 & 0.699 & 0.712\\
            Single Task  & \textsc{ FCN} \cite{wang2017time} & 0.631  & 0.791 & 0.806 & 0.871\\
               & \textsc{1D-FF} & 0.484 & 0.726 & 0.845 & 0.947 \\ \hline
               
            No covariates     & \textsc{N-BEATS} \cite{oreshkin2020meta} & 0.193 & 0.677 & 0.924  & $\times$ \\
            
                 &  & $\pm$~ 0.002 & $\pm$~ 0.005 & $\pm$~ 0.006 & $\times$ \\
               & \textsc{SCINet} \cite{liu2021time} & 0.192 & 0.594 & 0.718 & $\times$ \\
               
              &  & $\pm$~ 0.003 & $\pm$~ 0.006 & $\pm$~ 0.006 & $\times$ \\ \hline
                
            Oracle     & Best Channel*  &  0.353 & 0.353 & 0.353 & 0.353\\ \hline
        \end{tabular}

    \end{table}     \subsection{Baseline methods}
        We evaluated our approach against baselines from different related problem settings since there is no approach that can learn across multivariate time-series tasks with heterogeneous channels to the best of our knowledge: a set of heuristics consisting of predicting the constant zero (\textsc{zero prediction}) as our data is normalized to mean zero, predicting the last observed time step of the target channel (\textsc{last time step}), and predicting the average of the last time steps over the covariate channels (\textsc{avg time step}). Moreover, we evaluate a set of models on each individual task in meta-test, namely a stacked GRU network (\textsc{gru}), a fully-convolutional model (\textsc{fcn}) \cite{wang2017time} and a feed-forward neural network which relies only on the last observed time step (\textsc{1d-ff}). Note that these models have a lower model complexity than the other approaches, as they are only trained on the support data of a single task without incorporating any other tasks. Additionally, we show the performance of a hypothetical oracle which gives the mean squared error between the target and the last time step of the closest covariate channel (\textsc{oracle}). This model can be used as a point of orientation for the upper limit of the information within the covariate channels and is not a feasible model.
        
        Moreover, we evaluate a set of time-series forecasting models by training them across all datasets without using the covariate channels. Here, we select only the target channel $y$ and the future target $y'$ per task, similar to the setup of Oreshkin et al. \cite{oreshkin2020meta}. In this setup we evaluate the current best model for time-series forecasting on \textit{PapersWithCode}, \textsc{SCINet}, and the popular time-forecasting model \textsc{N-BEATS} \cite{nbeats} as it is shown as a successful model for for zero-shot time-series forecasting on univariate data \cite{oreshkin2020meta}. Evaluating the state-of-the-art forecasting models on the single-task level has proven to be infeasible, as a single task is too small for these architectures to train on, as well as too computationally expensive for all tasks in the test set.
        Lastly, we evaluate \iwata{} by feeding the last time-steps of $x'$ and $y$ of the time-series as it is the only few-shot learning model which can learn with heterogeneous attributes.
        The details on all baseline approaches and the training setup can be found in the supplementary material. 
    \begin{figure}[t!] \label{abl1} \centering
            \includegraphics[width=0.8\textwidth]{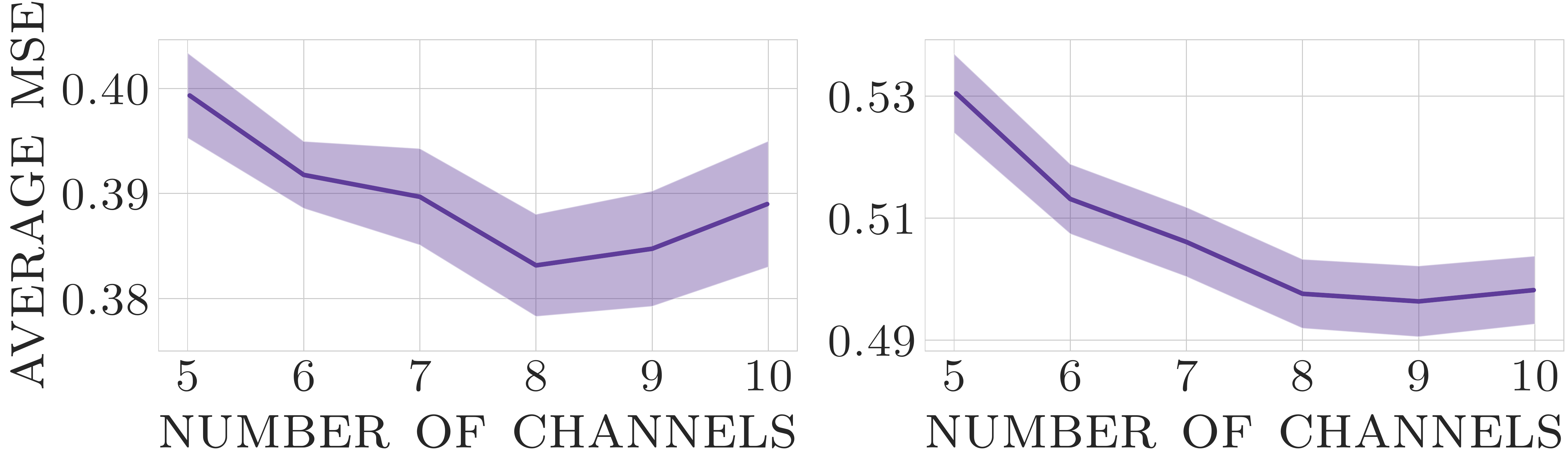}
            \caption{\textbf{Varying number of channels:} Mean-squared error over the test tasks with fixed number of channels. Results are averaged over all five folds and five repetitions with the corresponding standard deviation (shade) for the experiment $t_{0}+10$ (left) and $t_{0}+80$ (right). } 
    \end{figure} 
    \subsection{Results}
    The results for our main experiments can be seen in Table \ref{resultsmain} stating the mean-squared error averaged over the test tasks of each fold. It is not useful to compute the standard deviation over cross-validation folds as each fold includes different datasets for which the expected losses naturally vary. Instead, we repeated the full 5-fold cross-validation experiment five times to compute a standard deviation. However, this was too computationally expensive for the models trained on each task from scratch. Thus it is only given for the approaches that train across tasks. Note that heuristical approaches have no standard deviation since they are completely deterministic, and the test tasks are pre-generated. This is the same reason why \textsc{zero prediction}, \textsc{avg time step} and \textsc{oracle} have the same performance for all experiments, as the test tasks only vary in what part of the target channel is given.
    
    Our approach is shown to outperform all of our baselines in all 4 scenarios. As expected, \textsc{last time step} yields competitive results for the $t_{0}+1$ experiments since oftentimes the next time step does not deviate too much. The closest baseline is \iwata{} which learns across tasks and utilizes the heterogeneous channels but does not incorporate any past temporal information. The approach is still significantly worse than our proposed model for $t_{0}+1$, $t_{0}+10$ and $t_{0}+80$ when looking at the standard deviation, suggesting that especially existing temporal information for the target channel aids the prediction.
    The models trained on a single task from scratch show an expected subpar performance as they can only learn from a single limited few-shot task. Moreover, the state-of-the-art time-series forecasting methods show a good performance on $t_{0}+1$ and $t_{0}+10$, but degrade a lot for $t_{0}+80$, which is the consequence of only relying on the target channel, while especially for $t_{0}+80$ the covariates are shown to be crucial. This fact is emphasized when evaluating the performance of \textsc{oracle} as it gives an upper limit on the achievable score when only considering the predictor channels. Note that there are no scores listed for approaches that rely solely on the target channel for $t_{0}+100$ as there is no past target $y$ given in that setting.

    \subsection{Ablations}
    \begin{figure}[t!] \label{abl2} \centering
            \includegraphics[trim={1cm 0.23cm 0.2cm 0.11cm},clip,width=0.72\textwidth]{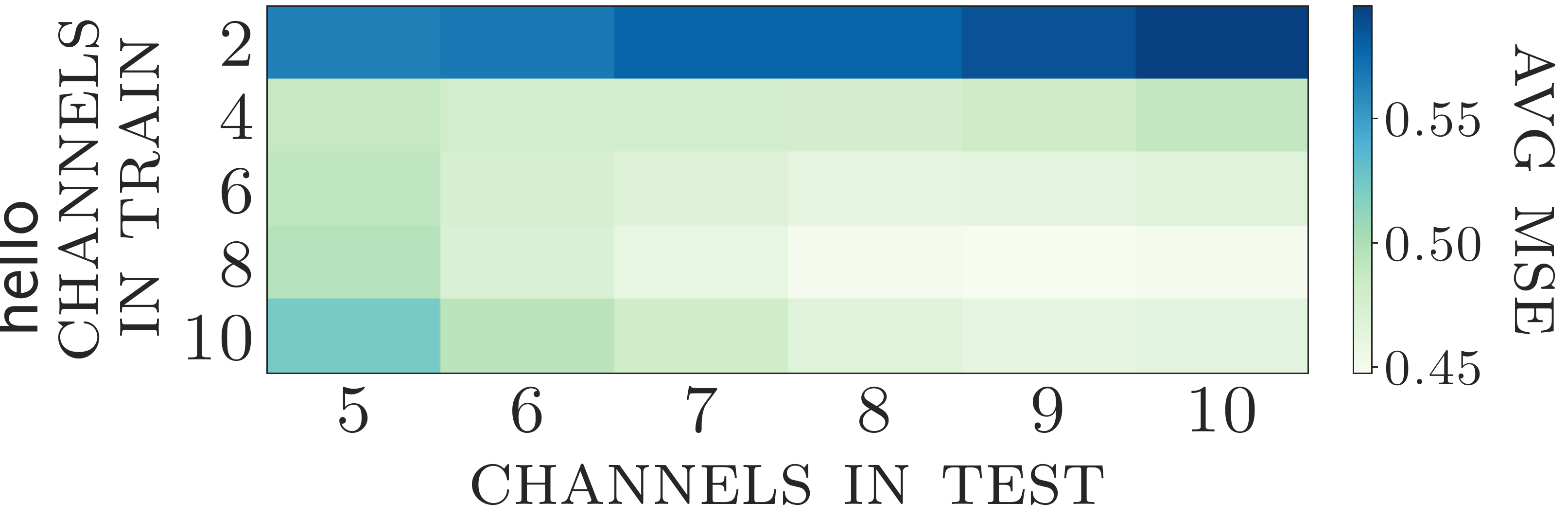}
            \caption{\textbf{Heatmap over channels:} Performance of our approach for $t_0+80$ when trained and evaluated on a fixed task size. Results are averaged over five repetitions of the 5-fold cross validation. Train channels include [2,4,6,8,10].}
    \end{figure}
    We conducted several ablation studies to analyze the robustness of our model with respect to the number of channels across tasks and to show the performance of different architectural choices.
    First, we analyzed the performance of our approach for $t_{0}+10$ and $t_{0}+80$ with respect to the number of predictor channels for a given novel test task. We show the aggregated results for this experiment in Figure \ref{abl1}. We can see that the performance of our approach increases with the number of predictor channels up to $8$ channels. Note that $c$ channels here refer to one target channel and $c-1$ predictor channels for each task. This suggests that our model successfully learned to process tasks independently of their number of channels. It can benefit from the fact that tasks with a higher number of randomly sampled channels are more likely to feature covariates that correlate to the target channel up to a certain point. In the case of predicting $10$ steps ahead, the performance slightly degrades for tasks with more than $8$ channels.

    To investigate the robustness of our model to the number of predictor channels of the tasks in training, we repeated our experiments while limiting the tasks sampled in training to a fixed number of channels. In Figure \ref{abl2} we show the performance of our model for $t_{0}+80$ trained only on tasks with $2$, $4$, $6$, $8$ and $10$ channels respectively, while evaluating it again for a fixed number of channels in test. One can see that the model is generally robust to the number of channels in training, with the exception of training on tasks with only $2$ channels, meaning only one predictor and one target channel. This is most likely due to the number of sampled training tasks that will not have a predictor channel that is sufficiently correlated to the target task. In accord with the previous ablation, there seems to be a slight performance optimum around tasks with $8$ channels, while training on tasks with $10$ channels and evaluating on tasks with $2$ channels also degrades the performance.
    
    Finally, we compare the performance of our model when changing the network design within the deep set blocks. Namely, we evaluate different combinations of convolutional and GRU layers to the one chosen in this work after optimizing the hyperparameters on the validation data. The results are shown in Table \ref{tab:results}. The best architecture which is used throughout all our experiments utilizes stacked GRU layers in first the deep-set blocks formalized in Equation \ref{bar} which receive the raw data input as well as the one before the final output layer in equation \ref{zlayer} which output the final embedding, while using 1D-convolutional layers in the intermediate blocks (\textit{GRU-Corner}). Using only GRU layers (\textit{All GRU}) degrades the performance by a slight margin while using convolutional layers at the beginning and end of the network is shown to be significantly worse (\textit{Conv-Corner}). This indicates that the GRU blocks are adapting more easily to the very heterogeneous time-series tasks.
    For comparison reasons, we make our code available at \url{https://github.com/radrumond/timehetnet}.

        \begin{table}[t!] 
        \centering
         \caption{\textbf{Architectures ablation:} Comparison of four different architecture setups.}
        \label{archtable}
        \begin{tabular}{ccccccccc}
            \hline
Experiment&~ & GRU-Corner & ~ & Conv-Corner & ~ & All GRU & ~ & All Conv\\\hline
        $t_{0}+80$&~ &\textbf{0.509} & ~ & 0.538       & ~ & 0.512   & ~ & 0.522   \\\hline
        $t_{0}+10$&~ &\textbf{0.389} & ~ & 0.397       & ~ & \textbf{0.389}   & ~ & 0.395   \\\hline
        \end{tabular}
       
        \label{tab:results}
    \end{table}  
    
\section{Conclusion}

In this work, we presented the first multivariate time-series forecasting model that works across tasks with heterogeneous channels. Currently, to the best of our knowledge, this is the first work to build a multivariate time-series meta-dataset for this type of meta task. Our model significantly outperforms all related baselines, which either fail to incorporate covariate information or cannot learn across tasks.
This approach serves as a benchmark for future research in this area. 
In future work, we would like to explore the effects of different deep-set blocks and how the model behaves with different types of models and problems. 
\subsubsection{Acknowledgements} This work was supported by the Federal Ministry for Economic Affairs and Climate Action (BMWK), Germany, within the framework of the IIP-Ecosphere project (project number: 01MK20006D).

\bibliographystyle{splncs04}

\appendix
\newpage
\section{Extra Experimental Details}

We have implemented both \iwata{} and \netname{} using Tensorflow 2.7. Each experiment ran on a machine equipped with a GPU card NVIDIA 1080ti. The average run-time was around 5 hours per experiment.
    
    \subsection{Dataset}
        The meta dataset used in our research is described in Table \ref{mtable}.
            \begin{table}[b]
    \centering
        \caption{Datasets used in this paper. Line with an empty `samples' value means they are one large single time-series. Keep in mind that the number of channels reported on this table correspond only to numerical information provided by the original sources.}
    \begin{tabular}{l|l|l|l|l} 
    Dataset                   & Source   & Samples & Time Stamps    & Channels \\ \hline
ArticularyWordRecognition & UCR      & 575   & 144    & 9       \\
    AtrialFibrillation        & UCR      & 30    & 640    & 2       \\
    BasicMotions              & UCR      & 80    & 100    & 6       \\
    Cricket                   & UCR      & 180   & 1197   & 6       \\
    DuckDuckGeese             & UCR      & 100   & 270    & 1345    \\
    EigenWorms                & UCR      & 259   & 17984  & 6       \\
    Epilepsy                  & UCR      & 275   & 206    & 3       \\
    ERing                     & UCR      & 300   & 65     & 4       \\
    EthanolConcentration      & UCR      & 524   & 1751   & 3       \\
    FaceDetection             & UCR      & 9414  & 62     & 144     \\
    FingerMovements           & UCR      & 416   & 50     & 28      \\
    HandMovementDirection     & UCR      & 234   & 400    & 10      \\
JapaneseVowels            & UCR      & 640   & 29     & 12      \\
    Libras                    & UCR      & 360   & 45     & 2       \\
    LSST                      & UCR      & 4925  & 36     & 6       \\
    MotorImagery              & UCR      & 378   & 3000   & 64      \\
    NATOPS                    & UCR      & 360   & 51     & 24      \\
    PEMS-SF                   & UCR      & 440   & 144    & 963     \\
    PhonemeSpectra            & UCR      & 6668  & 217    & 11      \\
    SelfRegulationSCP1        & UCR      & 561   & 896    & 6       \\
    SelfRegulationSCP2        & UCR      & 380   & 1152   & 7       \\
    SpokenArabicDigits        & UCR      & 8798  & 93     & 13      \\
    StandWalkJump             & UCR      & 27    & 2500   & 4       \\
    UWaveGestureLibrary       & UCR      & 440   & 315    & 3       \\
    ETTh1                     & Informer &         & 17420   & 7        \\
    ETTh2                     & Informer &         & 17420   & 7        \\
    ETTm1                     & Informer &         & 69680   & 7        \\
    ETTm2                     & Informer &         & 69680   & 7        \\
    WTH                       & Informer &         & 35064   & 12       \\
    ECL                       & Informer &         & 26304 & 321     \\
    cnc                       & Kaggle   & 18      & 2099    & 46       \\
    mining                    & Kaggle   &         & 737453 & 23       \\
    plant\_monitoring         & Kaggle   & 10      & 31841   & 25       \\
    peekdb                    & Peek     & 55      & 7151    & 20       \\
    Covid                     & Monash   &         & 212     & 266      \\
    Electricity               & Monash   &         & 26304   & 321      \\
    FRED                      & Monash   &         & 728     & 107      \\
    Temperature Rain          & Monash   & 422     & 725     & 76       \\
    Traffic                   & Monash   &         & 17544   & 862      \\
    Rideshare                 & Monash   & 156     & 541     & 15       \\ \hline
    \end{tabular}

    \label{mtable}
    \end{table}         All datasets are openly available at the time of writing.

    \subsection{Layer Blocks}
        For our implementation of \iwata{}, we kept the exact same structure as described in their paper.
        In the implementation of \netname{}, each layer block in Figure \ref{network} consists of 3 stacked temporal layers. We have selected two types of temporal layers, 1-dimensional convolutions, and GRUs.
        For our best configuration, layers representing the functions in equations \ref{bar} and \ref{zlayer} use GRU layers while the ones in \ref{ulayer} and \ref{vlayer} uses convolutions.
        For the convolutional layers, we have fixed the kernel size to 3 and channel output to 32 and used ReLU as activation functions. While for GRU layers, we fixed the number of cells to 32 and used Tanh as our activation.
        We use the last time-step of the output of the GRU from the last GRU layers in $g_{z}$ as our $z$ embedding. For the ablations where we used convolution layers in $g_{z}$, we applied a global average pooling following the convolution output.
        Layers in $f_{y},f_{c}$ and $g_c$ were defined as dense layers for our experiments.
        In order to concatenate the embeddings for $y$ with the ones of $[x,x']$, we padded $y$ with zeros.
        For more specific implementation details, please, refer to our code.
        
    \subsection{Specific Hyperparameters}
        We optimize the hyperparameters for \iwata{} and \netname{} via grid search. These are the hyperparameters we used throughout our main experiments:
        \begin{itemize}
            \item \iwata{}:
                \begin{itemize}
                    \item We fixed the latent dimensions for all layers to 32.
                    \item Adding gradient clipping targeting norm 1.0 was helpful for the $t+80$ experiments.
                \end{itemize}
                \item \netname{}:
                \begin{itemize}
                    \item For $t+1$ we've changed the inference layers' channel output dimension to 16 and 32 for the prediction layers. For the rest of them we used 32 for all.
                    \item Adding gradient clipping targeting norm 1.0 was helpful for the $t+1$ experiments.
                \end{itemize}
                \item Both:
                \begin{itemize}
                    \item Learning rate was set to 0.001
                    \item We trained for 15000 epochs
                    \item We have also used early stopping with a patience of 1500 epochs while saving the best weights.
\end{itemize}      
        \end{itemize}

        \subsection{Training setup for forecasting baselines}
        
        Similar to the setup in \cite{oreshkin2020meta}, we evaluated both \textsc{SCINet} and \textsc{N-BEATS} for zero-shot univariate time-series forecasting since both methods are not equipped to deal with heterogeneous covariates and offer no trivial way of adapting the model on the support samples of a new task without retraining the full model. Thus, both models were trained on the target channels $y$ of the tasks in meta-training with the objective of predicting $y'$, while we evaluated them for forecasting $y'$ based on $y$ for each query dataset $D^q$ in test.

\end{document}